\documentclass[12pt]{article}

\usepackage[table]{xcolor}
\usepackage{makecell}
\usepackage{enumerate}
\usepackage{color}
\usepackage{subfig}
\usepackage{caption}
\usepackage{amsfonts}
\usepackage{amsmath}
\usepackage{amssymb}
\usepackage{times}
\usepackage{cite}
\usepackage{hhline}
\usepackage{compactbib}
\usepackage{graphicx}        
\usepackage{amsmath,amsfonts,amssymb}
\usepackage{geometry}
\usepackage{lineno,hyperref}
\usepackage[labelsep=period]{caption}

\definecolor{halfgreen}{RGB}{0,128,0}
\definecolor{ahsred}{RGB}{192,0,0}

\renewcommand{\thefigure}{\arabic{figure}}
\renewcommand{\thetable}{\Roman{table}}

\usepackage{xcolor, soul}
\sethlcolor{yellow}
\usepackage{comment}

\usepackage{tikz}
\usetikzlibrary{shapes, arrows, positioning, fit, backgrounds}
\usetikzlibrary{intersections}

\newcommand{\etal}{\textit{et al.} }

\newcommand{\ie}{\textit{i}.\textit{e}. }

\usepackage{authblk}

\begin{document}

	\title{\bf When is an SHM problem a Multi-Task-Learning problem?}
        \author[1]{S.C.\ Bee}
	\author[2]{L.A.\ Bull}
	\author[1]{N.\ Dervilis}
 	\author[1]{K.\ Worden}
	\affil[1]{Dynamics Research Group, Department of Mechanical Engineering, University of Sheffield, Mappin Street, Sheffield S1 3JD, UK}
	\affil[2]{Department of Engineering, University of Cambridge, United Kingdom, UK, CB3 0FA}

	\date{}
	\maketitle
	\thispagestyle{empty}

%----------------------------------------------------------------------------------------------------------------------
% 1st PAGE
%----------------------------------------------------------------------------------------------------------------------

\vspace*{0mm}

%----------------------------------------------------------------------------------------------------------------------
\noindent \uppercase{\textbf{ABSTRACT}} \vspace{12pt} 

Multi-task neural networks learn tasks simultaneously to improve individual task performance. There are three mechanisms of multi-task learning (MTL) which are explored here for the context of structural health monitoring (SHM): (i) the natural occurrence of multiple tasks;  (ii) using outputs as inputs (both linked to the recent research in population-based SHM (PBSHM)); and, (iii) additional loss functions to provide different insights. Each of these problem settings for MTL is detailed and an example is given.

%----------------------------------------------------------------------------------------------------------------------

%------------------------------------------------------------------------------------------------------------
% SECTION 1: INTRODUCTION
%------------------------------------------------------------------------------------------------------------

\vspace{24pt} 
\noindent \uppercase{\textbf{INTRODUCTION}}  \vspace{12pt} 

Structural health monitoring (SHM), is a predictive tool which provides an online damage detection and condition monitoring strategy using data recorded from an individual structure \cite{Farrar2013}. Data are often unavailable or incomplete, measurements can be limited \cite{Bull2021} and, as a consequence, a data set which would be used to train a model could be insufficient to provide reliable results. Part of the increased motivation for SHM systems is from the growing number of structures which are reaching the end of their design life \cite{Farrar2013}; if the condition of the structures can be accurately assessed online for damage, then the structures can continue to operate. A key element to a SHM system is that it is \emph{accurate}, as the cost of uninformative predictions is not just economical but could have safety implications too.

Multi-task learning (MTL) refers to a suite of algorithms which learn tasks simultaneously, as opposed to in isolation from each other. MTL can be applied to improve generalisation of tasks, hence the accuracy of predictions, and therefore it could be applied to improve SHM systems. Caruana \cite{Caruana1997}, developed one of the first forms of multi-task learning, a neural network (NN) with back propagation, to train tasks simultaneously and improve generalisation between tasks. Since the original work, multi-task learners have been applied to a lot of different machine-learning algorithms, from support vector machines \cite{Shiao2012} to decision trees \cite{wang2008, chapelle2011}. Within this paper, multi-task learning will be discussed purely in relation to NNs. Improved generalisation may improve accuracy of predictions across multiple tasks and hence could be beneficial to SHM systems.  

The purpose of this paper is to provide a discussion of the applicability of the different types of multi-task learning problem settings with respect to the field of SHM. To distinguish between whether an NN is multi-task or not, a non-MTL NN will be referred to as an independent learner and a multi-task NN will be referred to as MTL. 

%------------------------------------------------------------------------------------------------------------
% SECTION 2: BACKGROUND
%------------------------------------------------------------------------------------------------------------

\vspace{24pt}

\noindent \uppercase{\textbf{BACKGROUND}} \vspace{12pt}

It is useful to start with the architecture of a simple NN, as shown in Figure \ref{fig:NN}. Data entered via the input layer, where each node represents an individual feature/measurement that is added to the network. The hidden layer represents a \emph{set} of latent features which are constructed from the measured features; there can be multiple hidden layers and hence multiple sets of latent features. Finally, the output layer generates the final output from the network. 

\tikzset{%
  every neuron/.style={
    circle,
    draw,
    minimum size=2em
  },
  neuron missing/.style={
    draw=none, 
    scale=3,
    text height=-0.25cm,
    execute at begin node=\color{black}$\hdots$
  },
}

\begin{figure}

\begin{minipage}{.5\textwidth}
\centering
% Traditional NN
    \begin{tikzpicture}[x=1cm, y=1cm, >=stealth]
    
     \foreach \m/\l [count=\x] in {1,2,3,missing,4}
      \node [every neuron/.try, neuron \m/.try] (input-\m) at (2.5+\x, 0) {};
    
     \foreach \m [count=\x] in {1,missing,2}
       \node [every neuron/.try, neuron \m/.try ] (hidden-\m) at (3+\x*1.25, 2) {};
    
     \foreach \m [count=\x] in {1}
      \node [every neuron/.try, neuron \m/.try ] (output-\m) at (4.5+\x,4) {};
    
     \foreach \l [count=\i] in {1,n}
      \node [] at (hidden-\i) {$L_\l$};
    
    \foreach \i in {1,...,4}
      \foreach \j in {1,...,2}
        \draw [->] (input-\i) -- (hidden-\j);
    
    \foreach \l [count=\i] in {1,2,3,n}
      \draw [<-] (input-\i) -- ++(0, -1)
        node [left, midway] {$I_\l$};
    
    \foreach \l [count=\i] in {1}
      \draw [->] (output-\i) -- ++(0, 1)
        node [left, midway] {$O_\l$};
    
    \foreach \i in {1,...,2}
      \foreach \j in {1}
        \draw [->] (hidden-\i) -- (output-\j);
    
    \foreach \l [count=\y from 0] in {Input, Hidden, Output}
      \node [align=center] at (2, \y*2) {\l \\ Layer};
    
    \end{tikzpicture}
\captionof{figure}{NN with single task.}
\label{fig:NN}  
\end{minipage}
\begin{minipage}{.5\textwidth}

\centering

% NN with multi outputs

    \begin{tikzpicture}[x=1cm, y=1cm, >=stealth]
    
     \foreach \m/\l [count=\x] in {1,2,3,missing,4}
      \node [every neuron/.try, neuron \m/.try] (input-\m) at (2.5+\x, 0) {};
    
     \foreach \m [count=\x] in {1,missing,2}
       \node [every neuron/.try, neuron \m/.try ] (hidden-\m) at (3+\x*1.25, 2) {};
    
     \foreach \m [count=\x] in {1,missing,2}
      \node [every neuron/.try, neuron \m/.try ] (output-\m) at (3.5+\x,4) {};
    
     \foreach \l [count=\i] in {1,n}
      \node [] at (hidden-\i) {$L_\l$};
    
    \foreach \i in {1,...,4}
      \foreach \j in {1,...,2}
        \draw [->] (input-\i) -- (hidden-\j);
    
    \foreach \l [count=\i] in {1,2,3,n}
      \draw [<-] (input-\i) -- ++(0, -1)
        node [left, midway] {$I_\l$};
    
    \foreach \l [count=\i] in {1,n}
      \draw [->] (output-\i) -- ++(0, 1)
        node [left, midway] {$O_\l$};
    
    \foreach \i in {1,...,2}
      \foreach \j in {1,...,2}
        \draw [->] (hidden-\i) -- (output-\j);
    
    \foreach \l [count=\y from 0] in {Input, Hidden, Output}
      \node [align=center] at (2, \y*2) {\l \\ Layer};
    
    \end{tikzpicture}

\captionof{figure}{NN with multiple tasks.}
\label{fig:MTL NN} 
\end{minipage}

\end{figure}

NNs have been applied within the context of SHM in varying applications. Elkordy \etal \cite{Elkordy1994} is an early example of the use of NNs in the context of SHM for damage detection. Manson \etal \cite{Manson2003} and Mustapha \etal \cite{Mustapha2007} each use a multi-layer perceptron NN to classify the location of damage, on an aircraft wing and isotropic plate, respectively. Since these early examples of the use of NNs in SHM, there has been a plethora of work utilising NNs, from using auto-associative NNs to detect damage before a crack is visibly seen \cite{Dervilis2014}, to using a convolutional NN to classify wind turbine tower vibration health states \cite{Khazaee2022}, a systematic review of the use of convolutional NNs in SHM is given in \cite{Sony2021}.

\begin{figure}
    \centering

\begin{tikzpicture}[x=1cm, y=1cm, >=stealth]
    
     \foreach \m/\l [count=\x] in {1,2,3,missing,4}
      \node [every neuron/.try, neuron \m/.try] (input-\m) at (5+\x, 0) {};
    
     \foreach \m [count=\x] in {1,missing,2}
       \node [every neuron/.try, neuron \m/.try ] (Shidden-\m) at (5.5+\x*1.25, 2) {};

    \foreach \m [count=\x] in {1,missing,2, 3, missing, 4}
       \node [every neuron/.try, neuron \m/.try ] (hidden-\m) at (3+\x*1.5, 4) {};
    
     \foreach \m [count=\x] in {1,missing,2}
      \node [every neuron/.try, neuron \m/.try ] (output1-\m) at (4+\x, 6) {};

    \foreach \m [count=\x] in {1,missing,2}
      \node [every neuron/.try, neuron \m/.try ] (output2-\m) at (8.5+\x, 6) {};
    
     % \foreach \l [count=\i] in {1,n, 1, n}
     %  \node [] at (hidden-\i) {$L_\l$};
    
    \foreach \i in {1,...,4}
      \foreach \j in {1,...,2}
        \draw [->] (input-\i) -- (Shidden-\j);
    
    \foreach \l [count=\i] in {1,2,3,n}
      \draw [<-] (input-\i) -- ++(0, -1)
        node [left, midway] {$I_\l$};
    
    \foreach \l [count=\i] in {1,m}
      \draw [->] (output1-\i) -- ++(0, 1)
        node [left, midway] {$O_\l$};

    \foreach \l [count=\i] in {{{m+1}},n}
      \draw [->] (output2-\i) -- ++(0, 1)
        node [left, midway] {$O_\l$};
    
    \foreach \i in {1,...,2}
      \foreach \j in {1,...,2}
        \draw [->] (hidden-\i) -- (output1-\j);

    \foreach \i in {3,...,4}
      \foreach \j in {1,...,2}
        \draw [->] (hidden-\i) -- (output2-\j);

    \foreach \i in {1,...,2}
      \foreach \j in {1,...,4}
        \draw [->] (Shidden-\i) -- (hidden-\j);
    
    \foreach \l [count=\y from 0] in {Input, Shared Hidden, Hidden, Output}
      \node [align=center] at (2, \y*2) {\l \\ Layer};

\end{tikzpicture}

    \caption{Example of a deep NN with both a shared hidden layer and  split hidden layer.}
    \label{fig:sharedMTLNN}
\end{figure}
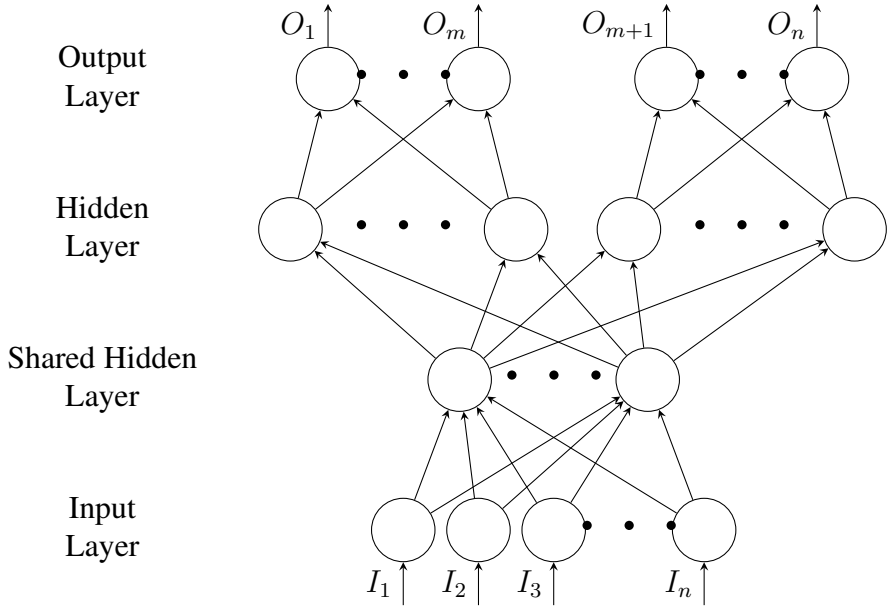

The structure of an MTL NN is an extension of the independent NN, as shown by Figure \ref{fig:MTL NN}. Just as with the independent formulations, the MTL NN has an input layer, hidden layers and an output layer. The main difference between Figure \ref{fig:NN} and Figure \ref{fig:MTL NN} is the additional output nodes in the output layer. The additional output nodes represent different tasks, \ie the multi-task nature of the network. The hidden layer could be more or less unchanged in the MTL setting, or it can take a very different structure. For a deep NN, the MTL NN may share all of the hidden layers or only several of the bottom hidden layers (an example of this structure is shown in figure \ref{fig:sharedMTLNN}). The input layer may have the same number of input features as with the independent learner; however, it may also have more input nodes to represent more features being added to the model. 

The structure of the NN will be dependent on the mechanism of MTL which is being implemented. There are three main themes of multi-task learning discussed here which have been inspired by Caruana's paper on MTL learning \cite{Caruana1997}, all which have applicability within SHM problem settings. Briefly these are:
\begin{itemize}
    \item Natural occurrence of multiple tasks - additional tasks which make sense to learn together.
    \item Using outputs as ``inputs'' - adding data which cannot be added as an input as an output so that the data can influence the model. This can be used when the data is not accessible on the same time scale as the other data inputs.
    \item Additional loss functions to provide different insights - repeating the output of a NN but using a different loss function.
\end{itemize}

The following sections will detail each of the mechanisms and provide insight into potential problem settings of them. 
\vspace{48pt}

%------------------------------------------------------------------------------------------------------------
% SECTION 3: Problem Setting 1: Natural occurrence of multiple tasks
%------------------------------------------------------------------------------------------------------------

\vspace{24pt}
\noindent \uppercase{\textbf{Problem Setting 1: Natural occurrence of multiple tasks}} \vspace{12pt}

Arguably, one of the most intuitive uses of MTL is when there is a \emph{set} of tasks which are related and therefore make sense to learn together. Often the definition of MTL is given in relation to this problem setting, Zhang \etal \cite{Zhang2018} provides an overview of multitask learning24the paper defines the aims of MTL as: \emph{to leverage useful information contained in multiple learning tasks to help learn a more accurate learner for each task}. For this problem setting of MTL, learning the tasks together provides a synergy that improves the performance of several tasks compared to if an independent learner was applied to each task individually. The structure of an MTL for this problem setting is given in Figure \ref{fig:sharedMTL1}.

\tikzstyle{input} = [rectangle, draw, fill=blue!20, 
    text width=6em, text centered, rounded corners, minimum height=4em]
\tikzstyle{hidden} = [rectangle, draw, fill=red!20, 
    text width=12em, text centered, rounded corners, minimum height=4em]
\tikzstyle{output} = [rectangle, draw, fill=green!20, 
    text width=6em, text centered, rounded corners, minimum height=4em] 
\tikzstyle{line} = [draw, -latex']

\begin{figure}[!h]
    \centering
\begin{tikzpicture}[node distance = 3cm, every node/.style={scale=0.8}]
    % Place nodes
    \node [input] (Input A) {Input A};
    \node [hidden, above right of=Input A] (Hidden) {Single Neural Network};
    \node [input, below right of =Hidden] (Input B) {Input B};
    \node [output, above right of=Hidden] (Output B) {Output B};
    \node [output, above left of=Hidden] (Output A) {Output A};
    
    % Draw edges
    \path [line] (Input A) -> (Hidden);
    \path [line] (Input B) -> (Hidden);
    \path [line] (Hidden) -> (Output A);
    \path [line] (Hidden) -> (Output B);

\end{tikzpicture}

    \caption{Problem setting 1: two tasks, A and B, put into one neural network.}
    \label{fig:sharedMTL1}
\end{figure}
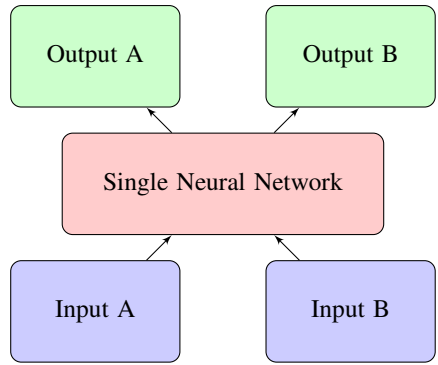

A example of the natural occurrence of multiple tasks in the context of SHM is the use of data from several nominally-identical structures to obtain information about their damage state, \emph{e.g.} from wind turbines in a wind farm. With multiple structures, generalisation may be improved such that physical changes may be more likely to be identified.

Obtaining information from damage states can be expensive; however, if data from different structures are combined, then damage-state information can be leveraged \emph{between structures} and provide increased information about the structures. One methodology of obtaining damage states for a structure is to synthesise it. Synthesised structures could be used within MTL to improve the performance of the actual structures which feature in the model. Synthesised data has the benefit of being significantly cheaper to obtain than the cost associated with providing damage to actual structures.

The majority of MTL which has been conducted in the field of SHM takes this form. In \cite{Li2021}, a multi-task Gaussian process regression is used for missing sensor data reconstruction across a dam sensor network. The data are not missing simultaneously; however, data points were missing from different sensors in the network and by looking at the data together, the reconstruction performance improved. In \cite{Huang2019}, multi-task sparse Bayesian learning was conducted on two damaged structures with supplementary information provided via simulations of the structures. Learning the two structures together, the damage patterns are more reliably detected. Finally, \cite{Lim2018} used an artificial neural network using data from six aluminium plates to predict fatigue crack length and remaining fatigue life. Although not explicitly called MTL in the paper, the neural network has two output tasks which are learnt jointly. 

A recent area of research is into population-based SHM (PBSHM) \cite{Bull2021, Gosliga2021, Gardner2021, Tsialiamanis2021, Bull2022}, which aims to utilise how data can be transferred and shared between \emph{populations} of structures to allow inferences to be shared across the population. An applicable example of a population of structures that may benefit from PBSHM is one of offshore wind farms. There now exist a lot of wind farms and hence a lot of structures which require SHM from a safety perspective, a cost perspective but also from an efficiency-of-power-generated perspective. This problem setting of MTL fits within the remit of PBSHM and could be explored within this context.

%------------------------------------------------------------------------------------------------------------
% SECTION 4: Problem Setting 2: Using outputs as inputs
%------------------------------------------------------------------------------------------------------------
\vspace{24pt}
\vspace{24pt}
\noindent \uppercase{\textbf{Problem Setting 2: Using outputs as inputs}} \vspace{12pt}

\begin{figure}[h]
    \centering
\begin{tikzpicture}[node distance = 3cm, every node/.style={scale=0.8}]
    % Place nodes
    \node [hidden] (Hidden) {Single Neural Network};
    \node [input, below of =Hidden] (Input A) {Input A};
    \node [input, above right of =Hidden] (Input B) {Input B};
    \node [output, above left of=Hidden] (Output A) {Output A};
    
    % Draw edges
    \path [line] (Input A) -> (Hidden);
    \path [line] (Hidden) -> (Output A);
    \path [line] (Hidden) -> (Input B);

\end{tikzpicture}

    \caption{Problem setting 2: Using additional inputs as an output.}
    \label{fig:sharedMTL2}
\end{figure}
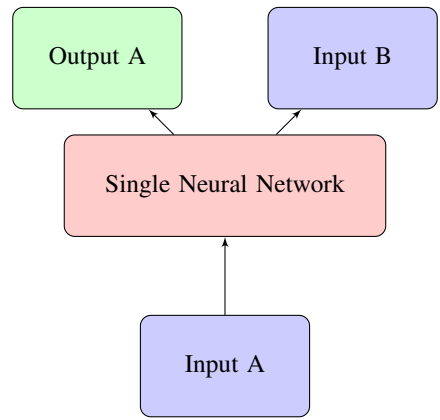

The second problem setting of MTL is using outputs as inputs, at first this may appear to be somewhat of an oxymoron; however, there are some features which are not available as inputs to a model, and therefore can only feature as outputs, see the structure in Figure \ref{fig:sharedMTL2}. An output influences a NN during back-propagation, which occurs during training, by having an impact on the weights within the model. Hence, although added as an output, the additional task will influence the model and can be viewed as a form of input. There are several different, but linked, forms that this can take: transfer learning, non-operational features, and regression for classification. Each of the different forms is expanded below.

Certain problem settings of MTL may be labelled as \emph{transfer learning}. Gardner \etal \cite{Gardner2022} broadly categorises transfer learning into two categories: training a model with data from an auxiliary task and fine tuning it with the main task data, and, performing domain adaptation such that two tasks can share a latent space. The latter category of transfer learning is applicable to SHM problem settings, with the aim to transfer knowledge between a \emph{source} domain and a \emph{target} domain \cite{Gardner2020}. Transfer learning does not require tasks to be learnt simultaneously; however, it is when tasks \emph{are} learnt simultaneously that transfer learning is also multi-task learning. 

An example of MTL transfer learning is given in \cite{Caruana1997}, in order to predict the medical risk of a patient, there are several features that can be measured as inputs (\emph{e.g.} height), however, there also may be medical test results which would be a useful feature to add to the model. Medical test results take time to process and may not be available as an input for all patients. Therefore, the test results can be added as an output to the model and form an auxiliary task. During training, the test results help to inform the medical risk. When tested, the target output is the risk and this can be calculated from the input features which are available. In this example information from the medical test results has been transferred to inform the medical risk. A potential synonymous example for SHM is using the results of model analysis as an additional output for a neural network which is used for system identification.

The examples above are also useful when considering the use of non-operational features during training. A further problem setting of using outputs as inputs to consider features that are available during training (which is likely offline, such as modal analysis), in comparison to features which are available during normal operation. The non-operational features that are available during training can be added as outputs to a NN which could improve the generalisation of the model and therefore improve the performance of the model during online utilisation. 

The final category for the problem setting of using outputs as inputs is using additional regression outputs to inform a classification task. For classification tasks, the classification is either True or False, and hence, there is a level of quantisation. Whereas regression tasks are on a sliding scale which can contain a lot more information than simply: True or False. Hence, there may be additional information in a regression task that would inform a classification task. It is a method of using a larger continuous space whilst solving a discrete problem. Additional classification tasks can also be used to improve the main classification task as the quantisation may be different to the original task. 

There is little research for this problem setting of MTL within the context of SHM. Aforementioned, there are strong links with transfer learning and there are promising problem settings of using outputs as inputs within the context of SHM.

%------------------------------------------------------------------------------------------------------------
% SECTION 4: Problem Setting 3: Additional loss functions to provide different insights
%------------------------------------------------------------------------------------------------------------

\vspace{24pt}
\noindent \uppercase{\textbf{Problem Setting 3: Additional loss functions to provide different insights}} \vspace{12pt}

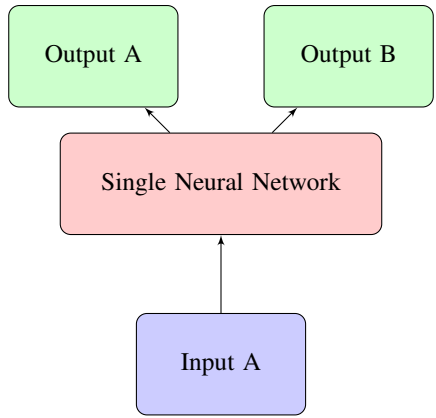
\begin{figure}[h]
    \centering
\begin{tikzpicture}[node distance = 3cm, every node/.style={scale=0.8}]
    % Place nodes
    \node [hidden] (Hidden) {Single Neural Network};
    \node [input, below of =Hidden] (Input A) {Input A};
    \node [output, above right of =Hidden] (Output B) {Output B};
    \node [output, above left of=Hidden] (Output A) {Output A};
    
    % Draw edges
    \path [line] (Input A) -> (Hidden);
    \path [line] (Hidden) -> (Output A);
    \path [line] (Hidden) -> (Output B);

\end{tikzpicture}

    \caption{Problem setting 3: Additional outputs for a NN.}
    \label{fig:sharedMTL3}
\end{figure}

When constructing a NN with a single task, the developer has to choose a loss function with which to train the network. During training, the weights and biases of the NN will be tuned to obtain the best performance of the NN, which is determined by minimising the loss function. Different loss functions may have affinities to different values of the weights and biases. Hence, the solution of the NN for different loss functions could provide different solutions and ould provide different insights from the network, even though the inputs into the network are exactly the same.

This problem setting of MTL duplicates outputs but uses different loss functions for each. Caruana \cite{Caruana1998} demonstrates how the rankprop error metric performs poorly at the lower end of a continuous spectrum. Following this, an MTL NN with both the rankprop error metric and sum-of-square errors metric is used to improve the performance of the main task at the lower end of the spectrum. Adding the additional loss function, in this case, improved performance at a critical end of a spectrum. 

A potential example of how this could be implemented within the field of SHM is in the context of fatigue testing. There are two parameters which would be of interest in fatigue testing: the number of cycles and the amplitude of the force experienced. To gauge the overall time signal, an L2-norm might be chosen; overall, it is anticipated that this would give a reasonable average response; however, it might not be sensitive to accurately detect the time when the loading switches from compression to tension. For this related task, a different regularisation parameter may be used, which focuses on when the response passes through 0. To understand the maximum load at the peak tension/compression, an L4-norm could be applied, which is useful at modelling extremities. As different regularisation may be used for different interpretations of the data, this is a good example of where MTL could improve the performance jointly over a range of tasks. 

This problem setting could have interesting applications within SHM. 

%------------------------------------------------------------------------------------------------------------
%                                          CONCLUDING REMARKS
%------------------------------------------------------------------------------------------------------------
\vspace{24pt}

\noindent \uppercase{\textbf{CONCLUDING REMARKS}} \vspace{12pt}

This paper discusses the different uses of MTL NNs and how then may be applicable in the field of SHM. MTL with multiple tasks arising naturally is the most explored of the problem settings, however, to date, there is still limited work using this approach. Transfer learning has been explored within PBSHM but the intersect of transfer learning and MTL is yet to be explored with regards to NNs. Arguably the least explored problem setting of MTL is the use of additional loss functions to provide different insights. As the cost of error in SHM can be quite high this problem setting could be very beneficial for improving prediction accuracy in the field. Overall there is a lot of potential for the use of MTL NNs in SHM, which can be researched.

\vspace{30pt}
\noindent \uppercase{\textbf{Acknowledgment}} \vspace{12pt}

This research was supported by grants from the Engineering and Physical Sciences Research Council (EPSRC), UK, and Natural Environment Research Council, UK via grant number, EP/S023763/1 and EP/W005816/1  - Revolutionising Operational Safety and Economy for High-value Infrastructure using Population-based SHM (ROSEHIPS).

%------------------------------------------------------------------------------------------------------------
%                                                 BIBLIOGRAPHY
%------------------------------------------------------------------------------------------------------------
\vspace{24pt}

\small 

\bibliographystyle{iwshm}
\bibliography{IWSHM_references}

%------------------------------------------------------------------------------------------------------------
%------------------------------------------------------------------------------------------------------------
%------------------------------------------------------------------------------------------------------------

\end{document}